\documentclass[letterpaper, 10 pt, conference]{ieeeconf}  

\IEEEoverridecommandlockouts                              

\usepackage{graphics} 
\usepackage{epsfig} 
\usepackage{times} 
\usepackage{amsmath}
\DeclareMathOperator{\Tr}{Tr}
\usepackage{amssymb}  
\usepackage{bm}
\usepackage{array}
\usepackage{float}
\usepackage{array}
\usepackage{duckuments}
\usepackage{todonotes}
\usepackage{booktabs}
\usepackage{multirow}
\usepackage{gensymb}
\usepackage{censor}
\usepackage{graphicx}
\usepackage{subcaption}

\DeclareMathOperator*{\argmax}{arg\,max}
\title{\LARGE \bf
Lend me an Ear: A Speech Enhancement Robotic Platform

}

\author{Zachary Turcotte and François Grondin
\thanks{This work was supported by the Natural Sciences and Engineering Research Council of Canada and the Canada Foundation of Innovation. Zachary Turcotte and François Grondin are with Department of Electrical and Computer Engineering,
         Université de Sherbrooke, Québec, Canada.
         {\tt\footnotesize zachary.turcotte@usherbrooke.ca, francois.grondin2@usherbrooke.ca}}%
}

\begin{document}

\maketitle
\thispagestyle{empty}
\pagestyle{empty}

\begin{abstract}

Speech enhancement performance degrades significantly in noisy environments, limiting the deployment of speech-controlled technologies in industrial settings, such as manufacturing plants. 
Existing speech enhancement solutions primarily rely on advanced digital signal processing techniques, deep learning methods, or complex software optimization approaches. 
This paper introduces a novel speech enhancement robotic platform that can reconfigure the geometry of a microphone array and adapt to changing acoustic conditions.
A sixteen-microphone array is mounted on a robotic arm manipulator with seven degrees of freedom.
The microphones are divided into four groups of four, including one group positioned near the end-effector.
The system reconfigures the array by adjusting the manipulator joint angles to place the end-effector microphones closer to the target speaker, thereby improving the reference signal quality. 
This proposed system is a multimodal sensing, reconfigurable audio capture device that integrates sound source localization techniques, computer vision, inverse kinematics, minimum variance distortionless response beamformer, and time-frequency masking using a deep neural network. 
Experimental results suggest that this approach outperforms other traditional recording configurations, achieving a higher average scale-invariant signal-to-distortion ratio and lower average word error rate across multiple input signal-to-noise ratio conditions. 

\end{abstract}

\section{INTRODUCTION}

The current labour shortage affects many manufacturing sectors in industrialized countries, and the integration of robots into the workforce has emerged as a potential solution.
To ease this transition, worker-robot interaction must be intuitive, with speech-based instructions representing a natural and efficient communication modality \cite{okuno_robot_2015,norda_evaluating_2024,bingol_performing_2020}.
Compared with alternative control interfaces, such as touchscreens or keyboard-based inputs, speech has been shown to be the most efficient human-robot interaction method \cite{norda_evaluating_2024,kadri_llm-driven_2025}. 
For instance, Kadri et al. \cite{kadri_llm-driven_2025} demonstrated successful speech-based control of an industrial manipulator.
When combined with a vision module and a large language model-driven agent, the system achieved a high success rate in understanding instructions and executing them. 
However, reliable operation of such systems requires clean speech signals.
In industrial environments, microphones capture both speech and background noise, which significantly degrades the performance of automatic speech recognition (ASR) systems. 

Speech enhancement (SE) methods can generally be categorized into single-channel and multichannel approaches. Multichannel techniques are especially advantageous because they can leverage spatial information obtained from multiple microphones. 
By exploiting spatial cues, these methods typically achieve improved noise reduction and lower signal distortion, which in turn enhances the reliability of ASR systems. 

Most multichannel SE approaches rely on beamforming techniques, and modern methods can be classified into three main categories: 1) Hybrid approaches use a Deep Neural Network (DNN) to estimate statistical properties of the noisy signal, such as an ideal ratio mask (IRM), which is then used to estimate the speech and noise spatial covariance matrices (SCMs). The estimated SCMs are subsequently employed to compute the coefficients of a spatial filter, such as the Minimum Variance Distortionless Response (MVDR) or Generalized Eigenvalue Decomposition (GEV) beamformer \cite{erdogan_improved_2016,heymann_neural_2016};
2) A DNN directly predicts the beamforming filter coefficients from the noisy input signal or its features \cite{xiao_deep_2016,li_neural_2016}.
The network learns a mapping between the observed noisy signals and the optimal spatial filter weights, with the goal of learning beamformers that outperform analytically derived solutions; 
3) End-to-end approaches map raw audio signals or noisy short-time Fourier transform (STFT) features to either clean signals or predicted text outputs \cite{pandey_tparn_2022} \cite{ochiai_unified_2017}. 
The network learns to transform noisy inputs into clean outputs without explicitly modeling intermediate beamforming steps.
These models are often trained jointly with language models, as ASR is typically the final stage of the processing pipeline.
However, the performance of these enhancement algorithms deteriorates in the presence of strong background noise, making it challenging to deploy voice-controlled technology in a manufacturing environment where noise is pervasive and continuous. 

The effectiveness of multichannel beamforming depends not only on the BF algorithm, but also on the reference channel. Microphones located at different positions may record different input SNR, reverberation level, and different acoustic transfer function. Most microphone arrays used in robotic applications have a fixed and compact geometry \cite{grondin_lightweight_2019, maheux_t-top_2022}, limiting the diversity of the signal at each microphone.

The selection of the reference channel is a crucial step when using beamforming approaches. It influences the choice of the acoustic transfer function used for enhancement by affecting the distortion and signal to noise ratio (SNR) of the output \cite{araki_comparison_2018,lawin-ore_reference_2012,zhang_study_2021,grimm_phase_2016}. The impact is even larger for wide microphone arrays. Choosing the closest microphone to the target as the reference channel is generally considered a simple and low-complexity option because it tends to improve the input SNR and the level of the target. However, proximity does not guarantee the optimal reference channel under all acoustic conditions.

In this work, we introduce a speech enhancement robotic platform that can actively reconfigure a microphone array mounted on an industrial manipulator. Rather than relying on a fixed sensing geometry, the proposed system uses the manipulator’s articulated joints to reposition the microphones according to the target location prior to audio capture. This allows for some microphones to be placed closer to the target and enables the selection of a higher-quality reference channel for the MVDR beamformer. By actively improving the acoustic observation before speech processing, the proposed approach could potentially improve speech enhancement performance in challenging noisy environments.

The paper is organized as follows. Section \ref{sec:Proposed method} presents the robotic platform. Section \ref{sec:Exp et res} assesses the performance of the system. Finally, Section \ref{sec: Discusion and conc} concludes this paper and discusses future work.
\section{Speech Enhancement Robotic Platform}
\label{sec:Proposed method}

\begin{figure}
    \centering
    \vspace{5mm}
    \includegraphics[width=1\linewidth]{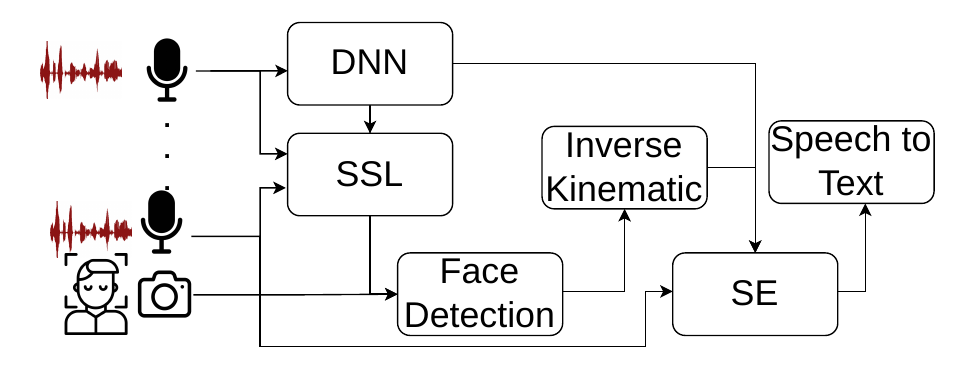}
    \caption{Reconfiguration algorithm}
    \vspace{-5mm}
    \label{fig:pipeline}
\end{figure}

The robotic platform reconfigures the microphone array to reposition microphones closer to the target speaker, improving speech capture and thereby speech enhancement and ASR performance. The proposed algorithm to reconfigure and denoise the speech signal is divided into five sequential modules, as illustrated in Fig. \ref{fig:pipeline}. 
The first module consists of a DNN that predicts IRMs. 
The second part is a {sound source localization} (SSL) module that estimates the approximate direction of arrival (DoA) of the target. 
Due to the presence of robotic arm components between the microphones and {the target sound source}, the free-field and far-field assumptions are violated, resulting in only approximate DoA estimation. 
Furthermore, the microphone positions vary with the arm configuration, preventing the use of methods based on fixed acoustic transfer functions, such as head-related transfer functions (HRTF). 
To refine target localization, the system leverages the integrated RGB camera and depth sensor. 
After obtaining an approximate azimuth angle from the SSL module, the robotic arm rotates so that the RGB camera and depth sensor are oriented toward the target. 
The third module uses the camera and depth sensor to localize the speaker in 3-D space using face detection. 
The fourth module consists of the inverse kinematic (IK) solver provided by the arm's manufacturer. 
The IK module receives the target localization and computes the joint angles required for the arm to reach a better listening position. 
The robotic arm subsequently repositions itself into the modified configuration. 
The fifth module applies an MVDR beamformer combined with a DNN-based IRM estimator.
This SE method is selected because it is inherently agnostic to the microphone array geometry. 

The main objective of this paper is to propose a flexible approach for integrating speech enhancement into noisy environments through the development of a robotic listening platform. The platform uses an articulated robotic arm to actively position microphones and improve speech capture under challenging noise conditions. By leveraging robotic hardware that can be incorporated into existing industrial workspaces, the proposed system enables effective voice interaction with robotic manipulators.

\subsection{Signal Model}
The signal can be modelled in the time-frequency domain as $\mathbf{Y}(t,f) = \mathbf{X}(t,f) + \mathbf{B}(t,f)$, where $\mathbf{Y}(t,f)\in \mathbb{C}^{K\times 1}$ stands for the Short-Time-Fourier-Transform (STFT) of the noisy recorded signals for $K \in \mathbb{N}$ microphones, $\mathbf{X}(t,f)\in \mathbb{C}^{K\times 1}$ for the STFT of the clean speech and $\mathbf{B}(t,f)\in \mathbb{C}^{K\times 1}$ for the STFT of the noise. Variables $t \in \{1, 2, \dots, T\}$ and $f \in \{0, 1, \dots, N/2\}$ stand for the time frame  index for a total of $T$ frames and the frequency bin with an $N$-sample analysis window, respectively.
\subsection{Ideal Ratio Masks estimation}
\label{subsec:DNN}
For both SE and SSL modules, it is essential to separate the target speech component $X_k(t,f) \in \mathbb{C}$ and the background noise component $B_k(t,f) \in \mathbb{C}$ from the noisy observation $Y_k(t,f) \in \mathbb{C}$ at each microphone $k$ to achieve reliable performance. This step corresponds to the DNN block of Fig. \ref{fig:pipeline}.
The IRM is an effective approach for performing this separation and is defined by:
\begin{equation}
    M_k(t,f) = \min \left( \max \left( \frac{|X_k(t,f)|}{|X_k(t,f)|+|B_k(t,f)|}, 0 \right), 1 \right),
    \label{eq:m_irm}
\end{equation}
where $M_k(t,f) \in [0,1]$ stands for the mask at frequency bin $f$, time frame $t$ and microphone $k$.
The operator $|\bm{\cdot}|$ denotes the magnitude of a complex value.
DNNs provide an effective way to estimate IRMs, and several network architectures have been proposed for this task.
In this study, a recurrent neural network (RNN) is used due to its relatively simple architecture and good performance. 
{The log magnitude spectrogram of the noisy signal ($\log|Y(t,f)|$) is fed as the input of the DNN.} 
The network begins with a batch normalization (BN) layer, followed by two bidirectional long short-term memory (BLSTM) layers with 1024 hidden units each.
The final layer is a fully connected (FC) layer with 257 output units, similar to the architecture proposed in SteerNet \cite{grondin_gev_2020}. 
The network is trained using single-channel recordings from the DNS Challenge dataset \cite{reddy_interspeech_2020}.
Reverberation is added using the SLR28 dataset, which contains both simulated and measured room impulse responses (RIRs) \cite{ko_study_2017}. 
During operation, the DNN processes the noisy input from all 16 channels and predicts 16 different masks $\{M_1(t,f),M_2(t,f),...,M_{16}(t,f) \}$.
\subsection{Sound Source Localization}
The Steered-Response-Power PHAse Transform (SRP-PHAT) algorithm with spatially whitened noise estimates the DoA of the target speaker \cite{grondin_lightweight_2019}.
This stage is illustrated by the SSL block of Fig. \ref{fig:pipeline}. 
The key difference from the conventional SRP-PHAT method lies in the use of the online Spatial Covariance Matrices (oSCMs) of both the noise ($\hat{\bm{\Phi}}_{B}(t,f) \in \mathbb{C}^{K \times K}$) and speech ($\hat{\bm{\Phi}}_{X}(t,f) \in \mathbb{C}^{K \times K}$) components:
\begin{equation}
    \hat{\bm{\Phi}}_{XX}(t,f) = (1-\alpha)\hat{\bm{\Phi}}_{XX}(t-1,f)+\alpha \hat{\bm{R}}_{XX}(t,f),
    \label{eq:oscn_s}
\end{equation}
\begin{equation}
    \hat{\bm{\Phi}}_{BB}(t,f) = (1-\alpha)\hat{\bm{\Phi}}_{BB}(t-1,f)+\alpha \hat{\bm{R}}_{BB}(t,f),
    \label{eq:oscm_b}
\end{equation}
where
\begin{equation}
    \hat{\mathbf{R}}_{XX}(t,f) = \hat{M}(t,f)\mathbf{Y}(t,f)\mathbf{Y}(t,f)^H,
\end{equation}
\begin{equation}
    \hat{\mathbf{R}}_{BB}(t,f) = (1-\hat{M}(t,f))\mathbf{Y}(t,f)\mathbf{Y}(t,f)^H,
\end{equation}
and $\alpha \in [0,1]$ is the adaptation rate, $\bm{\hat{R}}_{XX}(t,f)$ and $\bm{\hat{R}}_{BB}(t,f)$ are respectively the speech and noise cross spectra at time frame $t$ and frequency bin $f$.

The global mask $\hat{M}(t,f) \in \mathbb{R}$ is obtained as follows:
\begin{equation}
    \hat{M}(t,f) = \max(\{M_1(t,f), M_2(t,f), \dots, M_{16}(t,f)\})
\end{equation}
with each $M_k(t,f)$ corresponding to the DNN prediction for channel $k$, as described in  section \ref{subsec:DNN}. 
The whitened spatial covariance $\mathbf{P}(t,f)$ corresponds to:
\[
\begin{aligned}
\mathbf{P}(t,f)
&=
\begin{bmatrix}
\phi_{1,1}(t,f) & \cdots & \phi_{1,K}(t,f) \\
\vdots & \ddots & \vdots \\
\phi_{K,1}(t,f) & \cdots & \phi_{K,K}(t,f)
\end{bmatrix}, \\ \\
&=
\hat{\boldsymbol{\Phi}}_{BB}(t,f)^{-1}
\hat{\boldsymbol{\Phi}}_{XX}(t,f),
\end{aligned}
\]
where $\phi_{m,n}(t,f)$ is the element of the whitened spatial covariance matrix for each microphone pair. The direction of arrival $\bm{\theta}^*_t \in \mathcal{D} = \{v \in \mathbb{R}^3: \lVert v\rVert_2 = 1\}$ can be estimated for each time frame as the potential direction that maximizes the beamformer power:
\begin{equation}
    \bm{\theta}^*_t = \argmax_{\bm{\theta}}(E(\mathbf{P}(t,f),\bm{\theta})),
\end{equation}
and the power corresponds to:
\begin{equation}
    E(\mathbf{P}(t,f),\bm{\theta}) = \sum_{u=0}^{K}\sum_{v=0}^K\sum_{f=0}^{F}W_{u,v}(f,\bm{\theta})\phi_{u,v}(t,f),
\end{equation}
\begin{equation}
    W_{u,v}(f,\bm{\theta}) = \exp\left(j\frac{2\pi f}{N}\left[\frac{fs}{c}(\mathbf{r}_u - \mathbf{r}_v) \cdot \bm{\theta}\right] \right), 
\end{equation}
where $fs \in \mathbb{N}$ is the sample rate (samples/sec), $c \in \mathbb{R}^+$ is the speed of sound (343 m/s), $\bm{\theta} \in \mathcal{D}$ is the DoA in Cartesian coordinates and $\mathbf{r}_{k} \in \mathbb{R}^3$ is the microphone position in Cartesian coordinates.

\subsection{Face Detection}

Once the direction of arrival is estimated by the SSL module, the robotic arm rotates toward the target direction.
During this motion, the arm transitions into the face detection configuration, as illustrated in Fig. \ref{fig:gen3_with_sub_arrays}. This process is represented by the Face Detection block of Fig. \ref{fig:pipeline}.
The sixth joint is tilted at $80\degree$, which provides a wide vertical field of view (FOV) that enables the camera to detect individuals of varying heights. 

MediaPipe \cite{lugaresi_mediapipe_2019} and the RGB camera are used to locate the face of the target within the image frame. 
When combined with the depth sensor, a simple pinhole camera model \cite{khoshelham_accuracy_2012} and 3-D geometric reconstruction are employed to estimate the $(x,y,z)$ coordinates of the center of the face relative to the manipulator's reference axis. The RGB camera captures a single frame, while the depth sensor acquires multiple frames that are subsequently averaged to improve robustness.
\subsection{Inverse Kinematics}
The IK module, illustrated as the Inverse Kinematic block in Fig. \ref{fig:pipeline}, requires the desired position and orientation of the end-effector, along with an initial angular guess for each joint. 
The target end-effector position is provided by the output of the face-detection module, while its orientation is computed based on the current system state. 
Initial joint angle guesses are obtained by measuring the joint angles when the arm is in the desired configuration before operation; this initialization also biases the IK solver towards the desired solution and accelerates convergence. 
Once a valid solution is found, the manipulator is reconfigured to the new pose. 
Fig. \ref{fig:a_optim} illustrates the arm configuration when tracking a target at $45\degree$. 
%
%
\subsection{Speech Enhancement}

\label{subsec:SE}

Once the arm reaches its new configuration, the MVDR beamformer is applied to suppress noise and produce a cleaner signal.
The MVDR method minimizes the output signal variance while enforcing a unit gain constraint on a reference microphone, thereby preventing signal distortion.
The corresponding filter weights are computed using the following equation \cite{souden_optimal_2010}:
\begin{equation}
    \mathbf{w}_{mvdr}(f) = \frac{\bm{\Phi}_{BB}^{-1}(f)\bm{\Phi}_{XX}(f)}
    {\Tr(\bm{\Phi}_{BB}(f)^{-1} \bm{\Phi}_{XX}(f))}
    \mathbf{u},
    \label{eq:mvdr}
\end{equation}
where $\Tr(\cdot)$ denotes the trace operator and $\mathbf{u} \in \{0,1\}^K$ is a one-hot vector used to select the reference channel.
The speech and noise SCMs, $\bm{\Phi}_{XX}$ and $\bm{\Phi}_{BB}$, are computed using the predicted mask $M_k(t,f)$ for each channel $k$:
\begin{equation}
    \hat{X_k}(t,f) = Y_k(t,f) M_k(t,f),
\end{equation}
\begin{equation}
    \hat{B_k}(t,f) = Y_k(t,f) (1-M_k(t,f)),
\end{equation}
\begin{equation}
    \hat{\mathbf{X}}(t,f) = \left[
    \begin{array}{ccc}
    \hat{X}_1(t,f) & \dots & \hat{X}_K(t,f) \\
    \end{array}
    \right]^T
\end{equation}
\begin{equation}
    \hat{\mathbf{B}}(t,f) = \left[
    \begin{array}{ccc}
    \hat{B}_1(t,f) & \dots & \hat{B}_K(t,f) \\
    \end{array}
    \right]^T
\end{equation}
\begin{equation}
    \bm{\Phi}_{XX}(f) = \sum_{t=0}^{T}\mathbf{\hat{X}}(t,f)\mathbf{\hat{X}}(t,f)^H,
    \label{eq:phi_xx}
\end{equation}
\begin{equation}
    \bm{\Phi}_{BB}(f) = \sum_{t=0}^{T}\mathbf{\hat{B}}(t,f)\mathbf{\hat{B}}(t,f)^H.
    \label{eq:phi_bb}
\end{equation}
For SE, each mask is used individually (as opposed to SSL where they are pooled in a single mask) to provide better enhancement performance.

For SE, the complete SCM is calculated non-recursively because speech recognition can be performed offline, while for SSL the localization is estimated on a frame-by-frame basis, which requires the use of an online SCM. 

\section{Experimental setup and results}
\label{sec:Exp et res}
\subsection{Experimental Setup}
\begin{figure}
    \centering
    \includegraphics[width=0.9\linewidth]{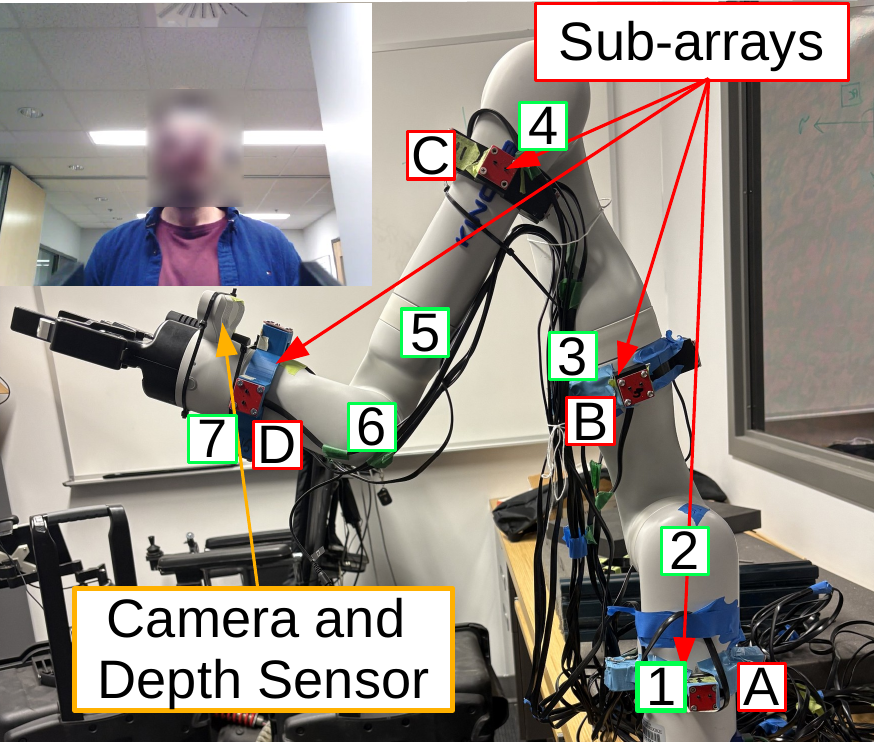}
    \caption{Experimental setup. Red markers indicate the sub-arrays location and green markers indicate the joint number.}
    \vspace{-5mm}
    \label{fig:gen3_with_sub_arrays}
\end{figure}

The proposed speech enhancement robotic platform consists of a Kinova Gen3 robotic arm with seven degrees of freedom and equipped with 16 omnidirectional microphones. 
These microphones are separated into four sub-arrays, A, B, C, and D, positioned on different joints of the robotic arm, including one sub-array located near the end-effector. 
The microphones are connected to the 16SoundsUSB sound card.
Fig. \ref{fig:gen3_with_sub_arrays} illustrates the precise placement of the sub-arrays and the robotic platform. Red boxes indicate the sub-array locations and green boxes indicate the joint numbers. 
The Gen3 is also equipped with an Omnivision 5640 RGB camera and an Intel RealSense Depth Module D410. We inspected the spectrograms of the captured signals and did not observe any ego-noise artifacts from the arm actuators while the arm is static.

The primary objective of the following experiments is to evaluate the speech-processing capabilities of the proposed robotic platform. The experiments assess whether the platform can accurately localize a speaker and enhance the captured speech under controlled conditions with degraded acoustics.
The experimental evaluation is organized into three sections. First, the performance of the SSL method is assessed, as localization constitutes a key component of the proposed system. Second, the proposed reconfigurable system is compared with conventional static microphone arrays and a shotgun microphone to quantify the benefits of dynamically positioning the microphones. Finally, the complete robotic platform is evaluated to determine its ability to localize the target, configure the microphone array, and improve speech enhancement performance in a controlled environment.
In all subsequent experiments, a printed image of a human face is mounted on a Genelec 8020D loudspeaker to approximate the visual appearance of a human speaker. Although this setup does not fully reproduce a real human subject, it provides a consistent visual reference that allows the face-detection model to operate reliably. This choice isolates the evaluation of SE from the additional variability introduced by natural human head motion and spontaneous speech, which we identify as the next stage of validation for future work.

\begin{figure}
    \centering
    
    \begin{subfigure}[t]{0.15\textwidth}
        \centering
        \includegraphics[width=\linewidth]{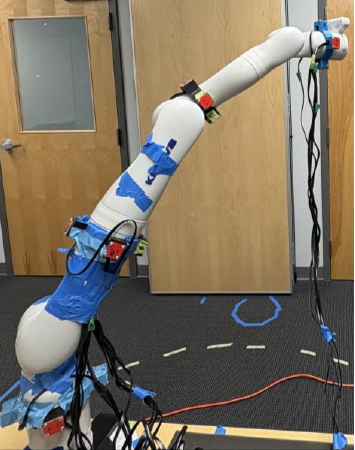}
        \caption{Reconfigured Array}
        \label{fig:a_optim}
    \end{subfigure}
    \hfill
    \begin{subfigure}[t]{0.15\textwidth}
        \centering
        \includegraphics[width=\linewidth]{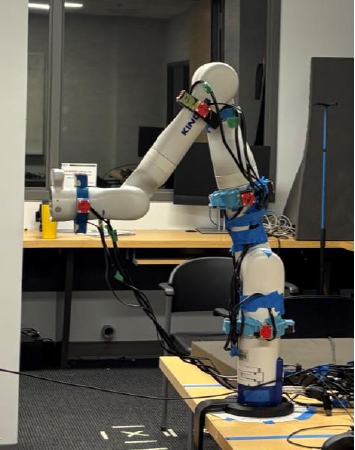}
        \caption{Static Array 1}
        \label{fig:b_stat1}
    \end{subfigure}
    \hfill
    \begin{subfigure}[t]{0.15\textwidth}
        \centering
        \includegraphics[width=\linewidth]{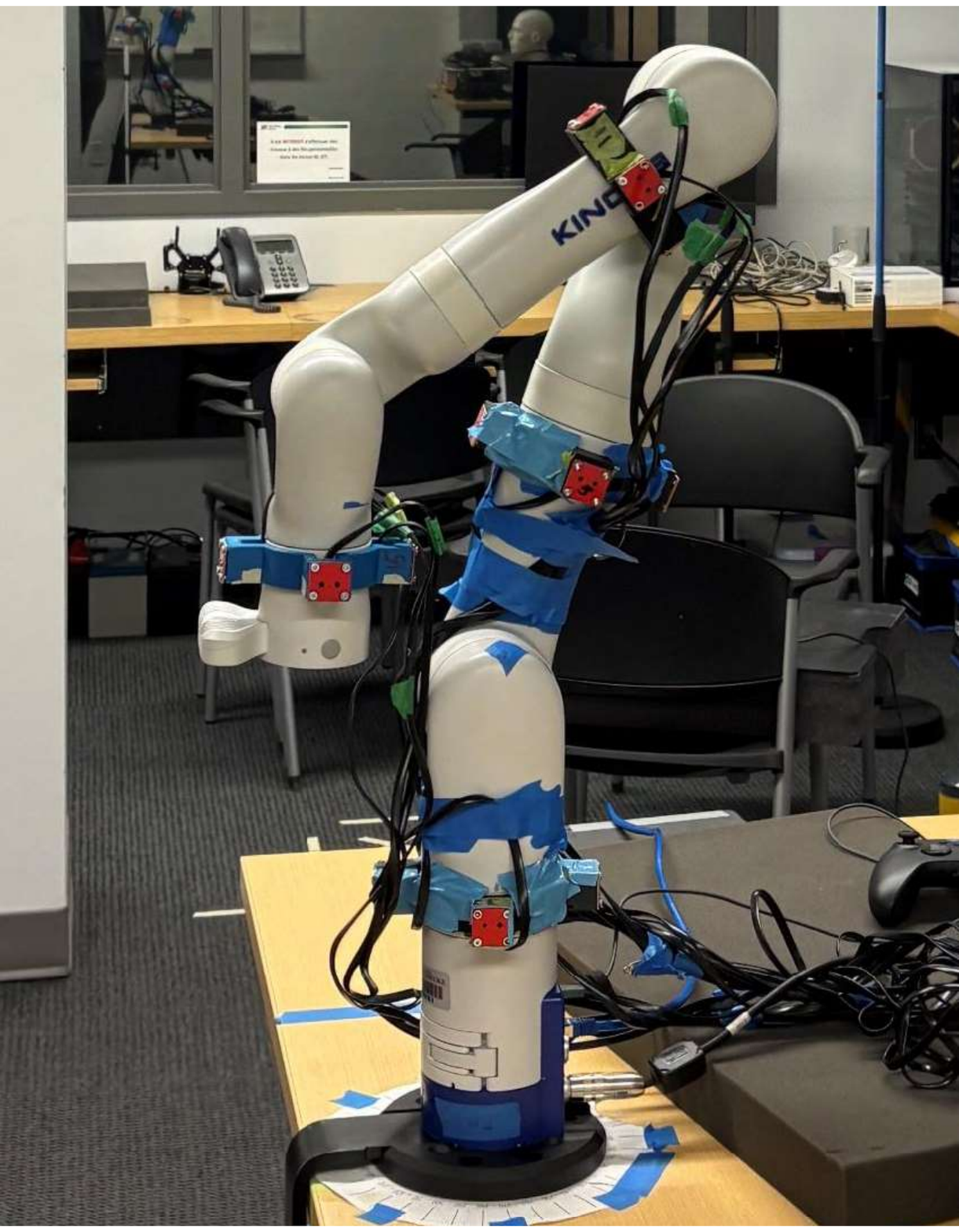}
        \caption{Static Array 2}
        \label{fig:c_stat2}
    \end{subfigure}

    \vspace{0.1cm}

    \begin{subfigure}[t]{0.15\textwidth}
        \centering
        \includegraphics[width=\linewidth]{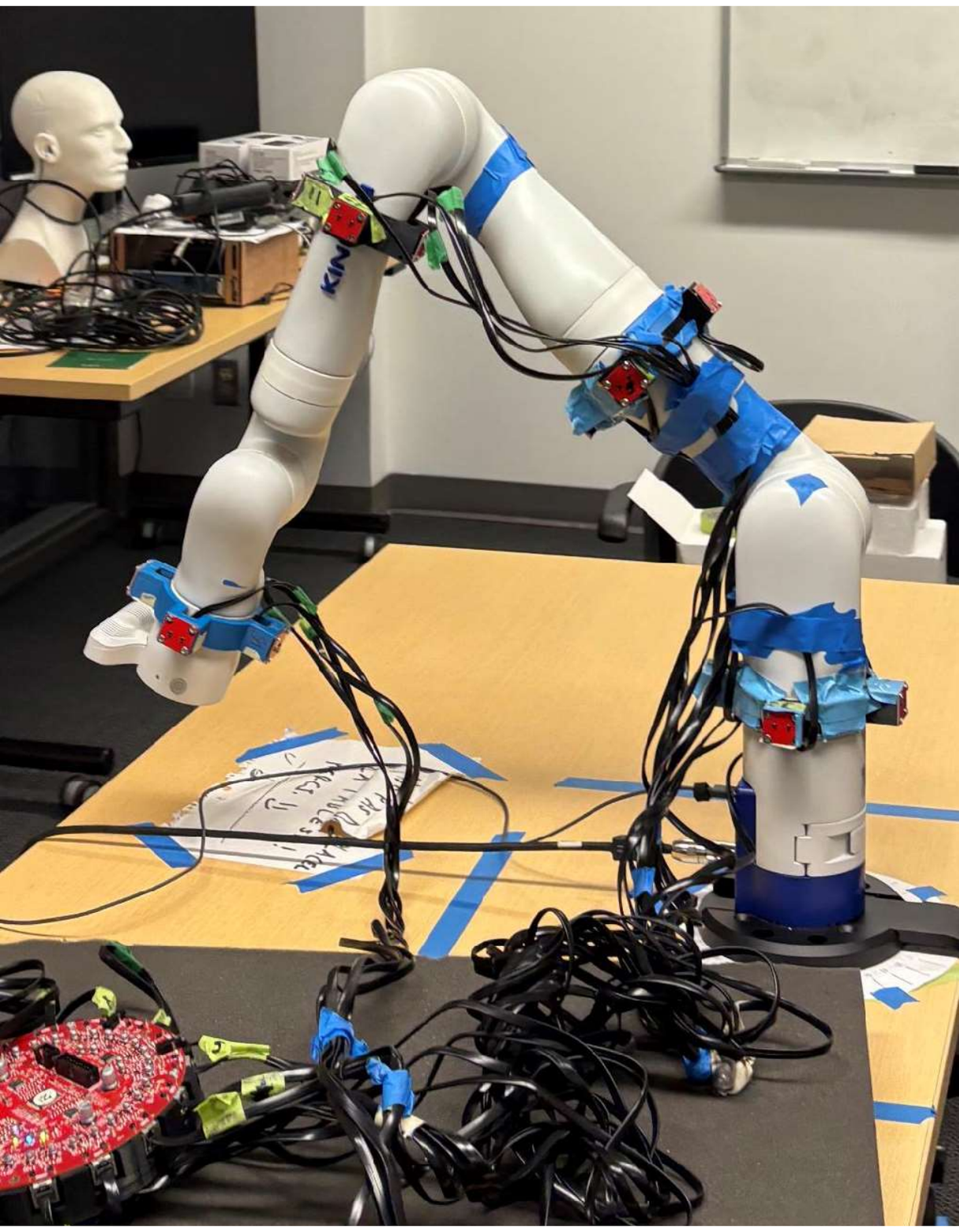}
        \caption{Static Array 3}
        \label{fig:d_stat3}
    \end{subfigure}
    \hfill
    \begin{subfigure}[t]{0.15\textwidth}
        \centering
        \includegraphics[width=\linewidth]{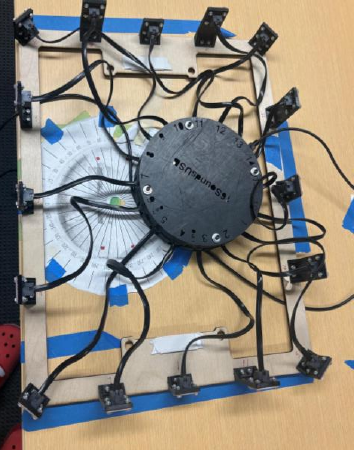}
        \caption{Square Array}
        \label{fig:e_square}
    \end{subfigure}
    \hfill
    \begin{subfigure}[t]{0.15\textwidth}
        \centering
        \includegraphics[width=\linewidth]{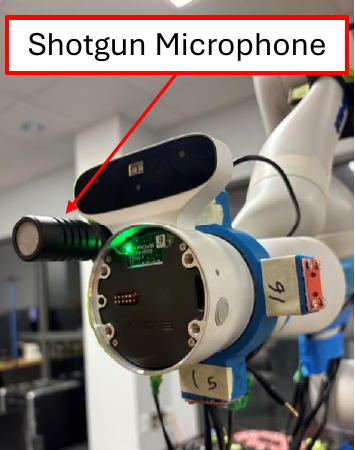}
        \caption{Shotgun Mic}
        \label{fig:f_shotgun}
    \end{subfigure}
    \caption{All recording devices used during experiments}
    \label{fig:all_arrays}
\end{figure}
\subsection{Sound Source Localization}
For the SSL evaluation, the free-field and far-field assumptions no longer hold when using all 16 microphones. The microphone array aperture can reach up to 1 m, while the expected interaction distance {varies between 1.2 and 2 m}. 
Because the system operates in the near-field, the plane-wave assumption of the classical SRP-PHAT formulation breaks down across most of the speech frequency band, degrading DoA estimation accuracy. In addition, parts of the robotic arm are located between some microphones and the target, causing acoustic wave diffraction, which violates the free-field assumption.
To better satisfy the far-field assumption, only sub-arrays A and B are used for computation. The distance between {these sub-arrays} is approximately {0.25 m}, resulting in a smaller array aperture. At the expected interaction distances, this configuration provides a reasonable far-field approximation over most of the relevant speech frequency range.
This configuration offers a good compromise between computational complexity and localization performance while satisfying one of the two fundamental assumptions of the conventional SRP-PHAT method.

To evaluate the SSL performance of the system, a three-second speech segment is recorded at 18 different DoAs $(0\degree,20\degree,40\degree,...,340\degree)$.
In addition, six different noise types (vacuum pump, drill, engine, electric noise, compressed air, mine noise) are recorded at 18 different DoAs $(10\degree,30\degree,50\degree,...,350\degree)$. 
Each speech recording is mixed with each noise recording at four different SNR levels $(-5,0,5,10)$ dB for a total of 1,944 mixtures per SNR level. 
Since SSL is the first processing step, the arm configuration would theoretically be random. Therefore, only one arm configuration is evaluated.
Fig. \ref{fig:b_stat1} shows the tested configuration. For evaluation, the selected output is the median azimuth $\bm{\theta}^*_t$ computed over all time frames $t$. The adaptation rate $\alpha$ is set to $0.1$ when calculating the oSCM.

The SSL performance is evaluated using the error rate with a $\pm15\degree$ tolerance ($ER_{15}$). 
The camera has a horizontal field of view of approximately $40\degree$.
Empirical evaluation shows that, with a $±15\degree$ localization error, the system can still reliably locate faces when the manipulator is correctly oriented.
Table \ref{tab:res_ssl} presents the $ER_{15}$ results and suggests that the localization performance remains acceptable at lower SNRs, with an $ER_{15}$ of 80\% using the NN mask.
As expected, the oracle mask outperforms the DNN-based mask at low SNR.
However, as the input SNR increases, the success rate improves, and the performance of the DNN IRM gradually approaches that of the oracle IRM.
\begin{table}[!ht]
    \centering
    \caption{$ER_{15}$ with oracle IRM and DNN IRM}
    \begin{tabular}{ccc}
    \toprule
       SNR (dB)  &  Oracle $ER_{15}\ (\%)$ & DNN $ER_{15}\ (\%)$ \\
       \midrule
        -5 & 89.45 &  79.89 \\
        0 & 93.78 & 90.43 \\
        5 & 96.81 & 95.42 \\
        10 & 98.41 & 98.20 \\
        \bottomrule
    \end{tabular}
    \label{tab:res_ssl}
\end{table}

\subsection{Speech Enhancement}

\label{subsec:SE}

For the first part of the SE evaluation, the reference channel will be the 16th, which is included in sub-array D. In most frameworks, the reference channel is less critical since no obstacle blocks the propagation path between the microphones and the array aperture is limited. Hence, the signals received at each microphone are similar. 
In the proposed system, however, the aperture can reach up to 1 m and sub-array D is positioned closer to the mouth of the target.
This placement enables the capture of a higher-quality speech reference and provides more reliable information to the MVDR beamformer. 

 The enhancement performance is compared against four different static microphone arrays and a small shotgun microphone. By varying only the audio capture device, this comparison allows us to assess whether the robotic platform outperforms conventional microphone configurations.
Fig. \ref{fig:all_arrays} presents the audio capture devices considered in the evaluation. 
The first three static arrays correspond to the microphones mounted on the robotic arm in a fixed configuration.
The fourth static array is a square array, the same used in \cite{lagace_ego-noise_2023}, and demonstrates the performance of a more conventional array, not directly mounted on the robot joints. 
Finally, a small shotgun microphone is mounted near the end-effector of the arm. 
Enhancement performance is evaluated using two metrics: 1) the Scale-Invariant Signal-to-Distortion Ratio (SI-SDR) \cite{roux_sdr_2019}; and 2) the word error rate (WER). 
Speech recognition is performed using the base Whisper model \cite{radford_robust_2022}. 
Both metrics are computed using the oracle and DNN IRMs. 
For each recording device, speech is recorded from four different DoAs ($45\degree,135\degree,225\degree,315\degree$), while noise is recorded from 18 DoAs $(10\degree,30\degree,50\degree,...,350\degree) $. 
It should be noted that, for recordings made with the reconfigured array and the shotgun microphone, the arm is reconfigured as a function of the speech DoA, so the end-effector faces the speaker. 
Four different speech segments and six different noise types (vacuum pump, drill, engine, electric noise, compressed air, mine noise) are recorded. 
Each speech segment from the four DoAs is mixed with each noise segment from all noise DoAs at four SNR levels $(-5,0,5,10)$ dB, resulting in a total of 10,368 mixtures per SNR level. 

To ensure a fair comparison between the audio capture devices, a reference gain is computed using the first four microphones of the reconfigured array.
This gain is used to normalize the noise such that the input SNR is set to $(-5,0,5,10)$ dB. 
For a given test configuration (e.g., speech from $45\degree$ and drill noise from $70 \degree$), the same noise gain is applied across all recording devices. 
This approach makes it possible to account for geometric differences between the devices and for the way sound is captured by the various microphone arrays. 

\begin{table*}[t]
    \centering
    \caption{SI-SDR (mean ± STD, dB), pooled across  acoustic conditions.}
    \label{tab:sisdr_snr}

    \small
    \renewcommand{\arraystretch}{1.12}

    \begin{tabular*}{\textwidth}
        {@{\extracolsep{\fill}}cc|cccccc@{}}
        \toprule
        \textbf{Mask}
        & \textbf{SNR (dB)}
        & \textbf{Reconfig.}
        & \textbf{Stat. 1}
        & \textbf{Stat. 2}
        & \textbf{Stat. 3}
        & \textbf{Square}
        & \textbf{Shotgun} \\
        \midrule

        \multirow{4}{*}{\rotatebox[origin=c]{90}{Oracle}}
        & $-5$
        & \textbf{19.31 $\pm$ 1.78}
        & 16.58 $\pm$ 1.66
        & 16.65 $\pm$ 1.72
        & 16.93 $\pm$ 1.79
        & 17.12 $\pm$ 1.94
        & 0.11 $\pm$ 2.42 \\

        & 0
        & \textbf{23.22 $\pm$ 1.86}
        & 20.41 $\pm$ 1.72
        & 20.47 $\pm$ 1.80
        & 20.89 $\pm$ 1.87
        & 21.08 $\pm$ 2.00
        & 3.67 $\pm$ 2.42 \\

        & 5
        & \textbf{27.07 $\pm$ 1.95}
        & 24.13 $\pm$ 1.83
        & 24.24 $\pm$ 1.92
        & 24.81 $\pm$ 1.99
        & 24.96 $\pm$ 2.10
        & 7.72 $\pm$ 2.44 \\

        & 10
        & \textbf{30.90 $\pm$ 2.03}
        & 27.86 $\pm$ 1.94
        & 28.02 $\pm$ 2.04
        & 28.78 $\pm$ 2.10
        & 28.86 $\pm$ 2.20
        & 12.13 $\pm$ 2.45 \\

        \midrule

        \multirow{4}{*}{\rotatebox[origin=c]{90}{NN}}
        & $-5$
        & \textbf{18.51 $\pm$ 1.90}
        & 15.56 $\pm$ 2.05
        & 15.56 $\pm$ 2.06
        & 16.05 $\pm$ 2.10
        & 16.32 $\pm$ 2.35
        & 0.11 $\pm$ 2.42 \\

        & 0
        & \textbf{22.95 $\pm$ 1.88}
        & 20.44 $\pm$ 1.95
        & 20.31 $\pm$ 2.03
        & 20.71 $\pm$ 2.11
        & 20.94 $\pm$ 2.25
        & 3.67 $\pm$ 2.42 \\

        & 5
        & \textbf{27.09 $\pm$ 1.86}
        & 24.78 $\pm$ 1.95
        & 24.68 $\pm$ 2.05
        & 25.09 $\pm$ 2.17
        & 25.25 $\pm$ 2.25
        & 7.72 $\pm$ 2.44 \\

        & 10
        & \textbf{31.00 $\pm$ 1.90}
        & 28.81 $\pm$ 1.93
        & 28.83 $\pm$ 2.08
        & 29.36 $\pm$ 2.25
        & 29.43 $\pm$ 2.31
        & 12.13 $\pm$ 2.45 \\

        \bottomrule
    \end{tabular*}
\end{table*}
\begin{table*}[t]
    \centering
    \caption{WER (mean ± STD, \%), pooled across acoustic conditions.}
    \label{tab:overall_wer_snr}

    \small
    \renewcommand{\arraystretch}{1.10}

    \begin{tabular*}{\textwidth}
        {@{\extracolsep{\fill}}cc|cccccc@{}}
        \toprule
        \textbf{Mask}
        & \textbf{SNR (dB)}
        & \textbf{Reconfig.}
        & \textbf{Stat. 1}
        & \textbf{Stat. 2}
        & \textbf{Stat. 3}
        & \textbf{Square}
        & \textbf{Shotgun} \\
        \midrule

        \multirow{4}{*}{\rotatebox[origin=c]{90}{Oracle}}
        & $-5$
        & \textbf{10.03 $\pm$ 1.19}
        & 22.71 $\pm$ 3.55
        & 22.19 $\pm$ 3.70
        & 24.47 $\pm$ 3.93
        & 19.79 $\pm$ 3.87
        & 40.42 $\pm$ 7.18 \\

        & 0
        & \textbf{8.33 $\pm$ 0.54}
        & 16.53 $\pm$ 1.71
        & 15.70 $\pm$ 1.69
        & 17.67 $\pm$ 2.06
        & 13.16 $\pm$ 1.33
        & 25.24 $\pm$ 5.42 \\

        & 5
        & \textbf{7.86 $\pm$ 0.42}
        & 13.66 $\pm$ 0.86
        & 12.82 $\pm$ 0.74
        & 14.57 $\pm$ 0.86
        & 11.24 $\pm$ 0.54
        & 16.21 $\pm$ 3.01 \\

        & 10
        & \textbf{7.64 $\pm$ 0.28}
        & 12.73 $\pm$ 0.63
        & 11.88 $\pm$ 0.44
        & 13.34 $\pm$ 0.43
        & 10.69 $\pm$ 0.29
        & 11.39 $\pm$ 1.46 \\

        \midrule

        \multirow{4}{*}{\rotatebox[origin=c]{90}{NN}}
        & $-5$
        & \textbf{17.35 $\pm$ 5.76}
        & 36.80 $\pm$ 8.82
        & 34.77 $\pm$ 8.63
        & 39.12 $\pm$ 9.03
        & 36.13 $\pm$ 10.84
        & 40.42 $\pm$ 7.18 \\

        & 0
        & \textbf{10.47 $\pm$ 2.33}
        & 22.79 $\pm$ 5.15
        & 22.00 $\pm$ 5.07
        & 24.78 $\pm$ 5.87
        & 19.20 $\pm$ 5.64
        & 25.24 $\pm$ 5.42 \\

        & 5
        & \textbf{8.06 $\pm$ 0.63}
        & 15.60 $\pm$ 2.73
        & 15.01 $\pm$ 2.60
        & 16.73 $\pm$ 3.16
        & 12.45 $\pm$ 1.63
        & 16.21 $\pm$ 3.01 \\

        & 10
        & \textbf{7.37 $\pm$ 0.33}
        & 11.99 $\pm$ 1.28
        & 11.60 $\pm$ 1.21
        & 12.99 $\pm$ 1.21
        & 10.50 $\pm$ 0.48
        & 11.39 $\pm$ 1.46 \\

        \bottomrule
    \end{tabular*}
\end{table*}

The SI-SDR is computed as a weighted average and weighted standard deviation, with each segment weighted proportionally to its duration. A similar procedure is used for the WER, except that each segment is weighted according to its number of words. 
Table \ref{tab:sisdr_snr} reports the SI-SDR and Table \ref{tab:overall_wer_snr} reports the WER. Each table {presents} the results for each SNR level for each listening device type. The WER values are clipped to 100\% because in some cases, mostly at low SNR levels, the speech-to-text module produced more words than there are in the reference speech, resulting in WER values exceeding 100\%. In all subsequent tables, the results with the shotgun microphone are reported under both the oracle-mask and DNN-mask, although no mask is actually applied. This presentation is used to facilitate comparison between the microphone arrays and the shotgun microphone.

The results imply that the reconfigured array generally achieves higher SI-SDR and lower WER than the other configurations. {Table} \ref{tab:sisdr_snr} shows that the average SI-SDR of the reconfigured array is higher than the other listening devices, suggesting that its enhanced signals contained less combined residual noise, interference, and speech distortion. The improvement over the fixed-array configurations {varies} between 2 and 3 dB. This improvement was observed with both Oracle and NN masks. In terms of WER, the results suggest that the platform effectively improves speech recognition performance, especially at low SNR. For both mask types, at ($-5,0,5$) dB, the reconfigured array achieves approximately half the WER of the other audio capture configurations, corresponding to a relative WER reduction of about 50\%. The reconfigured array also generally exhibits lower standard deviations, indicating more consistent performance across the different acoustic conditions.

Additionally, a paired comparison was conducted to evaluate the WER obtained with the reconfigurable array against each recording device under identical acoustic conditions while using the 16th microphone as the reference channel. Each trial {is} matched according to the speech DoA, noise DoA, noise type, SNR, and speech segment. For instance, the WER values obtained for speech segment 1 at DoA of ($45^\circ$), a drill noise at DoA of ($190^\circ$) and the SNR set to -5 dB {are} compared directly across devices. For each matched trial, the WER of the reconfigurable array is compared to the WER of the other listening devices and classified as lower, equal, or higher. This comparison is used to determine how often the reconfigurable array achieves an equal or lower WER than the baseline configurations. A win indicates that the reconfigured array achieved a lower WER than the baseline under the same acoustic condition. The win rate {is} then computed as the proportion for which the reconfigurable array achieved a lower WER. 

Results for both mask types are compiled in {Table} \ref{tab:wer_win_comparison_combined}. The results suggest that in challenging acoustic conditions, the reconfigurable array {outperforms} other devices for both NN and oracle masks 80\% of the time. When considering the win-and-tie rate, it goes above 90\%. However, the win rate of the reconfigured array {drops} gradually as the SNR increases. At 10 dB, the win rate is between 50\% and 80\% for both mask types and the combined win-and-tie exceeds 80\% for all baselines, suggesting that the reconfigurable array either improves WER or achieves performance comparable to the other listening devices in the majority of evaluated conditions. It is interesting to note that at high SNR, the win rate of the reconfigured array compared to the shotgun microphone is 50\%, which indicates that the gain at high SNR could be minimal, but when compared to other arrays, the win rate stays between 60\% and 80\%.

To further validate that the reconfigured array provides the best performance, the same experiment was repeated, but the MVDR filter weights are computed using each channel in turn as the reference. 
For each configuration, the minimum WER obtained across all possible reference channels is selected, and shown in Table \ref{tab:min_wer_snr}. Results suggest that even if the reference yielding the best WER could be selected \emph{a priori}, the reconfigured array would still exhibit lower average WER than static arrays.

The overall speech enhancement results suggest that selecting the 16th microphone as the reference channel for the reconfigured array is a reasonable design choice: it surpasses the best-case performance of all static arrays at low SNR, although it does not strictly maximize performance for the reconfigured array itself. Additionally, when doing the matched trials the reconfigured array had a lower WER in more than 90\% of trials at low SNR when considering the win-and-tie. The results further indicate that the static arrays and shotgun microphone appear to perform similarly among the different test conditions. These findings imply that mounting the microphones directly on the arm without any geometry adjustment yields performance comparable to that of a conventional static array.

\begin{table}[t]
    \centering
    \caption{Pairwise WER comparison}
    \label{tab:wer_win_comparison_combined}

    \small
    \renewcommand{\arraystretch}{0.92}
    \setlength{\tabcolsep}{2.5pt}

    \resizebox{\columnwidth}{!}{%
    \begin{tabular}{cc|cc|cc|cc}
        \toprule
        \multirow{1}{*}{\textbf{SNR(dB)}}
        & \multirow{1}{*}{\textbf{Baseline}}
        & \multicolumn{2}{c|}{\textbf{Win (\%)}}
        & \multicolumn{2}{c|}{\textbf{Tie (\%)}}
        & \multicolumn{2}{c}{\textbf{Win+Tie (\%)}} \\

        
        & 
        & \textbf{Oracle}
        & \textbf{NN}
        & \textbf{Oracle}
        & \textbf{NN}
        & \textbf{Oracle}
        & \textbf{NN} \\

        \midrule

        \multirow{5}{*}{$-5$}
        & Static 1 & 85.01 & 88.72 & 6.83 & 4.86 & 91.84 & 93.58 \\
        & Static 2 & 84.95 & 89.05 & 6.66 & 4.52 & 91.61 & 93.57 \\
        & Static 3 & 88.48 & 91.78 & 4.80 & 3.76 & 93.28 & 95.54 \\
        & Square   & 82.12 & 87.33 & 8.10 & 5.56 & 90.22 & 92.89 \\
        & Shotgun  & 83.26 & 83.26 & 5.45 & 5.45 & 88.71 & 88.71 \\
        \midrule

        \multirow{5}{*}{$0$}
        & Static 1 & 80.84 & 84.43 & 8.45  & 7.06  & 89.29 & 91.49 \\
        & Static 2 & 80.03 & 84.49 & 10.07 & 6.42  & 90.10 & 90.91 \\
        & Static 3 & 86.28 & 89.47 & 6.94  & 4.98  & 93.22 & 94.45 \\
        & Square   & 74.88 & 79.75 & 10.71 & 7.75  & 85.59 & 87.50 \\
        & Shotgun  & 75.41    & 75.41 & 10.19    & 10.19 & 85.60    & 85.60 \\
        \midrule

        \multirow{5}{*}{$5$}
        & Static 1 & 75.06 & 77.37 & 11.69 & 11.86 & 86.75 & 89.23 \\
        & Static 2 & 78.07 & 80.03 & 10.24 & 9.95  & 88.31 & 89.98 \\
        & Static 3 & 83.97 & 85.76 & 6.71  & 7.00  & 90.68 & 92.76 \\
        & Square   & 68.17 & 72.51 & 13.48 & 11.05 & 81.65 & 83.56 \\
        & Shotgun  & 65.57    & 65.57 & 19.68    & 19.68 & 85.25 & 85.25 \\
        \midrule

        \multirow{5}{*}{$10$}
        & Static 1 & 71.41 & 66.67 & 15.74 & 18.58 & 87.15 & 85.25 \\
        & Static 2 & 75.58 & 71.47 & 12.79 & 15.45 & 88.37 & 86.92 \\
        & Static 3 & 80.56 & 78.59 & 9.03  & 10.42 & 89.59 & 89.01 \\
        & Square   & 64.76 & 67.30 & 15.10 & 14.58 & 79.86 & 81.88 \\
        & Shotgun  & 50.69 & 50.69 & 32.58 & 32.58 & 83.27 & 83.27 \\
        \bottomrule
    \end{tabular}%
    }
\end{table}

\begin{table*}[t]
    \centering
    \caption{Minimum WER (mean ± STD, \%) obtained using the best reference channel
    under various SNR conditions.}
    \label{tab:min_wer_snr}

    \small
    \renewcommand{\arraystretch}{1.12}

    \begin{tabular*}{\textwidth}
        {@{\extracolsep{\fill}}cc|ccccc@{}}
        \toprule
        \textbf{Mask}
        & \textbf{SNR (dB)}
        & \textbf{Reconfig.}
        & \textbf{Stat. 1}
        & \textbf{Stat. 2}
        & \textbf{Stat. 3}
        & \textbf{Square} \\
        \midrule

        \multirow{4}{*}{\rotatebox[origin=c]{90}{Oracle}}
        & $-5$
        & \textbf{7.07 $\pm$ 1.97}
        & 14.71 $\pm$ 3.96
        & 13.60 $\pm$ 4.02
        & 14.00 $\pm$ 3.60
        & 11.82 $\pm$ 3.49 \\

        & 0
        & \textbf{5.70 $\pm$ 1.00}
        & 9.86 $\pm$ 2.06
        & 9.12 $\pm$ 2.18
        & 9.28 $\pm$ 1.89
        & 7.65 $\pm$ 1.57 \\

        & 5
        & \textbf{5.35 $\pm$ 0.73}
        & 7.90 $\pm$ 1.39
        & 7.50 $\pm$ 1.44
        & 7.69 $\pm$ 1.43
        & 6.29 $\pm$ 1.10 \\

        & 10
        & \textbf{5.27 $\pm$ 0.64}
        & 7.30 $\pm$ 1.31
        & 6.96 $\pm$ 1.36
        & 7.16 $\pm$ 1.35
        & 5.84 $\pm$ 0.91 \\

        \midrule

        \multirow{4}{*}{\rotatebox[origin=c]{90}{NN}}
        & $-5$
        & \textbf{13.17 $\pm$ 5.84}
        & 25.59 $\pm$ 7.99
        & 23.38 $\pm$ 8.16
        & 25.54 $\pm$ 7.99
        & 24.40 $\pm$ 9.31 \\

        & 0
        & \textbf{7.48 $\pm$ 2.52}
        & 14.75 $\pm$ 4.82
        & 12.98 $\pm$ 4.60
        & 13.74 $\pm$ 4.36
        & 11.84 $\pm$ 4.43 \\

        & 5
        & \textbf{5.63 $\pm$ 0.98}
        & 9.33 $\pm$ 2.51
        & 8.26 $\pm$ 2.30
        & 8.76 $\pm$ 2.25
        & 7.05 $\pm$ 1.49 \\

        & 10
        & \textbf{5.17 $\pm$ 0.60}
        & 7.04 $\pm$ 1.41
        & 6.41 $\pm$ 1.29
        & 6.89 $\pm$ 1.33
        & 5.75 $\pm$ 0.77 \\

        \bottomrule
    \end{tabular*}
\end{table*}
%

\subsection{Robotic Platform Validation}

\begin{table}[!ht]
    \centering
     \caption{Different speaker configurations during testing}
    \begin{tabular}{ccccccc}
    \toprule
        Case & 1 & 2 & 3 & 4 & 5 & 6 \\
    \midrule
        Speech DoA & 30\degree & 60\degree & 90\degree & 120\degree & 140\degree & 170\degree \\ 
        Noise DoA & 270\degree & 240\degree & 340\degree & 270\degree & 20\degree & 300\degree \\
    \bottomrule
    \end{tabular}
    \label{tab:conf_integration}
\end{table}

The following experiment validates the entire robotic platform and assesses its feasibility. 
The first part of the experiment consists of simultaneously playing a five-second speech segment through a loudspeaker and a noise segment through another. 
During this period, the system records the signals and performs SSL. Once the azimuth angle is estimated, the robot rotates to face the target. 
Then, the face detection and IK modules estimate the face coordinates and move the end-effector close to the target. Finally, one loudspeaker plays speech, and another one plays noise; both signals are recorded independently for subsequent evaluation.
The complete system is evaluated under six different speech-noise DoA configurations, listed in Table \ref{tab:conf_integration}. 
The same four different speech segments and six noise types are used as defined in previous experiments.
These six configurations are also tested in real-world conditions at approximately the same four SNR levels $(-5,0,5,10)$ dB.

\begin{figure}[!ht]
    \centering
    \includegraphics[width=0.9\linewidth]{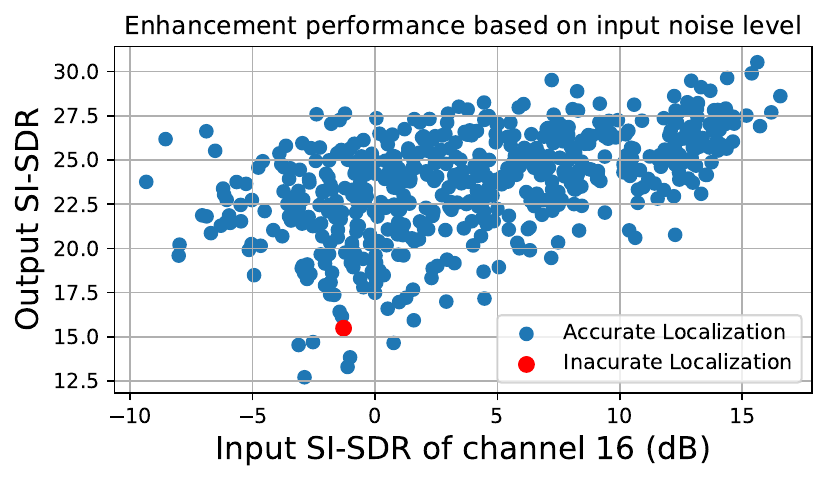}
    \caption{Output SI-SDR vs input SI-SNR at channel 16}
    \vspace{-10pt}
    \label{fig:si_sdr_pipeline}
\end{figure}

\begin{figure}[!ht]
    \centering
    \includegraphics[width=0.9\linewidth]{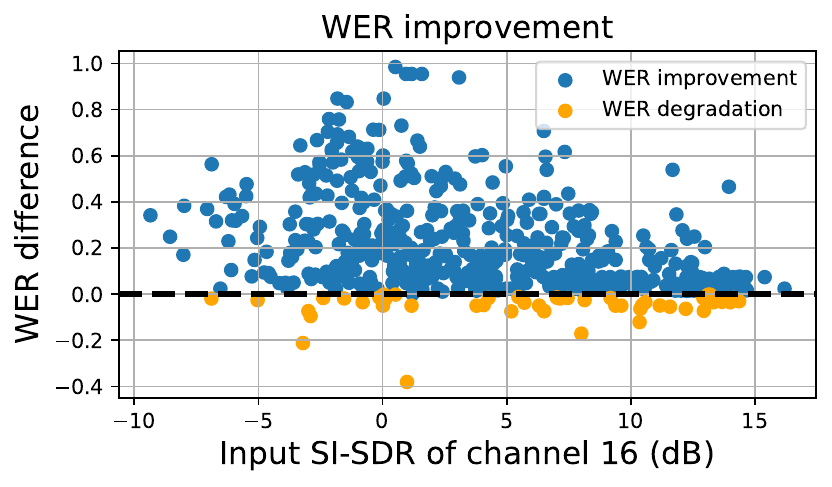}
    \caption{WER improvement vs the input SNR at channel 16}
    \vspace{-10pt}
    \label{fig:WER Improvement}
\end{figure}

Fig. \ref{fig:si_sdr_pipeline} illustrates the SI-SDR improvement as a function of the input SI-SDR of channel 16. It follows an upward trend, which was expected. 
During testing, the localization failed only once (red dot).
Fig. \ref{fig:WER Improvement} presents the WER difference relative to the unenhanced baseline, which corresponds to the unfiltered audio of channel 16.
The results indicate that speech enhancement has a more pronounced effect on the WER for low input SI-SDR recordings, whereas the magnitude of the improvement decreases as the input SI-SDR increases. This trend is expected since a signal with higher SI-SDR will naturally have lower WER. However, the figure also shows that the MVDR processing negatively affects the transcription output of several recordings. 
Finally, across all tested scenarios, the face detection and IK modules operated successfully, except when the initial SSL stage failed. However, these modules were not thoroughly evaluated, as their performance was not the primary focus of this experiment.

%

%
\section{Conclusion}
\label{sec: Discusion and conc}


This paper presents a speech enhancement robotic platform that reconfigures an arm-mounted microphone array to reposition microphones closer to the target speaker, improving speech capture and thereby speech enhancement and ASR in noisy environments.
The platform combines four main capabilities: SSL based on a noise-robust SRP-PHAT variant, face detection using MediaPipe, an inverse kinematics module, and an SE module combining an MVDR beamformer with a DNN-based time-frequency mask for speech-noise separation.

Experimental results assessing the SSL and SE capabilities indicate that the platform can efficiently locate a speaker in challenging acoustic conditions and, through a simple, low-complexity microphone reconfiguration scheme, achieve improved speech enhancement over conventional audio capture devices.
A separate experiment validated the full platform, providing evidence for the feasibility of the proposed approach in a noisy, controlled environment.

Future work will address the main limitations of this work. First, the current pipeline relies on accurate initial sound source localization to seed the downstream visual and kinematic stages, with no mechanism to recover using other modalities if this estimate is inaccurate. To address this, we plan to improve algorithmic robustness through tighter module coupling, for instance by incorporating visual target localization as a fallback when the initial SSL stage is inaccurate. Second, we plan to validate the platform's real-world effectiveness with real human subjects.

\section*{Acknowledgement}
Generative AI (ChatGPT and Claude) was used to improve text in this paper.

\bibliography{references}

@inproceedings{xiao_deep_2016,
	title      = {Deep beamforming networks for multi-channel speech recognition},
	pages      = {5745--5749},
	booktitle  = {Proc. of the IEEE ICASSP},
	author     = {Xiao, Xiong and Watanabe, Shinji and Erdogan, Hakan and Lu, Liang and Hershey, John and Seltzer, Michael L. and Chen, Guoguo and Zhang, Yu and Mandel, Michael and Yu, Dong},
    year = {2016},
}

@article{bingol_performing_2020,
	title      = {Performing predefined tasks using the human–robot interaction on speech recognition for an industrial robot},
	volume     = {95},
	pages      = {103903},
	journal    = {Eng. Appl. Artif. Intell.},
	author     = {Bingol, Mustafa Can and Aydogmus, Omur},
    year       = {2020},
}

@inproceedings{okuno_robot_2015,
	title      = {Robot audition: Its rise and perspectives},
	year       = {2015},
	pages      = {5610--5614},
	booktitle  = {Proc. of the IEEE ICASSP},
	author     = {Okuno, Hiroshi G. and Nakadai, Kazuhiro},
}

@article{kadri_llm-driven_2025,
	title      = {{LLM}-driven agent for speech-enabled control of industrial robots: A case study in snow-crab quality inspection},
	volume     = {27},
	pages      = {106660},
	journal    = {RINENG},
	author     = {Kadri, Ibrahim and Selouani, Sid Ahmed and Ghribi, Mohsen and Ghali, Rayen and Mekhoukh, Sabrina},
    year       = {2025},
}

@article{norda_evaluating_2024,
	title      = {Evaluating the Efficiency of Voice Control as Human Machine Interface in Production},
	volume     = {21},
	pages      = {4817--4828},
	number     = {3},
	journal    = {IEEE T-ASE},
	author     = {Norda, Marvin and Engel, Christoph and Rennies, Jan and Appell, Jens-E. and Lange, Sven Carsten and Hahn, Axel},
    year       = {2024},
}

@article{grondin_lightweight_2019,
	title      = {Lightweight and optimized sound source localization and tracking methods for open and closed microphone array configurations},
	volume     = {113},
    year       = {2019},
	pages      = {63--80},
	journal    = {Robot. Auton. Syst.},
	author     = {Grondin, François and Michaud, François},
}

@article{lugaresi_mediapipe_2019,
	title      = {{MediaPipe}: A Framework for Building Perception Pipelines},
	publisher  = {{arXiv}},
	author     = {Lugaresi, Camillo and Tang, Jiuqiang and Nash, Hadon and {McClanahan}, Chris and Uboweja, Esha and Hays, Michael and Zhang, Fan and Chang, Chuo-Ling and Yong, Ming Guang and Lee, Juhyun and Chang, Wan-Teh and Hua, Wei and Georg, Manfred and Grundmann, Matthias},
	journal    = {arXiv preprint {arXiv}:1906.08172},
    year       = {2019},
}

@article{souden_optimal_2010,
	title      = {On Optimal Frequency-Domain Multichannel Linear Filtering for Noise Reduction},
	volume     = {18},
	pages      = {260--276},
	number     = {2},
	journal    = {IEEE/ACM Trans. Audio Speech Lang. Process.},
	author     = {Souden, Mehrez and Benesty, Jacob and Affes, {SofiÈne}},
    year       = {2010},
}

@article{khoshelham_accuracy_2012,
	title      = {Accuracy and Resolution of Kinect Depth Data for Indoor Mapping Applications},
	volume     = {12},
	year       = {2012},
	pages      = {1437--1454},
	number     = {2},
	journal    = {Sensors (Basel, Switzerland)},
	author     = {Khoshelham, Kourosh and Elberink, Sander Oude},
}

@inproceedings{heymann_neural_2016,
	title      = {Neural network based spectral mask estimation for acoustic beamforming},
	year       = {2016},
	pages      = {196--200},
	booktitle  = {Proc. of the IEEE ICASSP},
	author     = {Heymann, Jahn and Drude, Lukas and Haeb-Umbach, Reinhold},
}

@inproceedings{erdogan_improved_2016,
	title      = {Improved {MVDR} Beamforming Using Single-Channel Mask Prediction Networks},
	year       = {2016},
	pages      = {1981--1985},
	booktitle  = {Proc. of Interspeech},
	author     = {Erdogan, Hakan and Hershey, John R. and Watanabe, Shinji and Mandel, Michael I. and Roux, Jonathan Le},
}

@inproceedings{maheux_t-top_2022,
	title      = {{T-Top}, a {SAR} Experimental Platform},
	year       = {2022},
	pages      = {904--908},
	booktitle  = {Proc. of the {ACM}/{IEEE} HRI},
	author     = {Maheux, Marc-Antoine and Caya, Charles and Létourneau, Dominic and Michaud, François},
}

@inproceedings{li_neural_2016,
	title      = {Neural Network Adaptive Beamforming for Robust Multichannel Speech Recognition},
    year       = {2016},
	pages      = {1976--1980},
	booktitle  = {Proc. of Interspeech},
	author     = {Li, Bo and Sainath, Tara N. and Weiss, Ron J. and Wilson, Kevin W. and Bacchiani, Michiel},
}

@inproceedings{pandey_tparn_2022,
	title      = {{TPARN}: {Triple}-Path Attentive Recurrent Network for Time-Domain Multichannel Speech Enhancement},
	year       = {2022},
	pages      = {6497--6501},
	booktitle  = {Proc. of the {IEEE} ICASSP},
	author     = {Pandey, Ashutosh and Xu, Buye and Kumar, Anurag and Donley, Jacob and Calamia, Paul and Wang, {DeLiang}},
}

@inproceedings{grondin_gev_2020,
	title      = {{GEV} Beamforming Supported by {DOA}-Based Masks Generated on Pairs of Microphones},
    year       = {2020},
    booktitle  = {Proc. of Interspeech},
	pages      = {3341--3345},
	author     = {Grondin, François and Lauzon, Jean-Samuel and Vincent, Jonathan and Michaud, François},
}

@inproceedings{reddy_interspeech_2020,
	title      = {The {INTERSPEECH} 2020 Deep Noise Suppression Challenge: Datasets, Subjective Testing Framework, and Challenge Results},
    booktitle  = {Proc. of Interspeech},
    year       = {2020},
	pages      = {2492--2496},
	author     = {Reddy, Chandan K. A. and Gopal, Vishak and Cutler, Ross and Beyrami, Ebrahim and Cheng, Roger and Dubey, Harishchandra and Matusevych, Sergiy and Aichner, Robert and Aazami, Ashkan and Braun, Sebastian and Rana, Puneet and Srinivasan, Sriram and Gehrke, Johannes},
}

@inproceedings{ko_study_2017,
	title      = {A study on data augmentation of reverberant speech for robust speech recognition},
	year       = {2017},
	pages      = {5220--5224},
	booktitle  = {Proc. of the {IEEE} ICASSP},
	author     = {Ko, Tom and Peddinti, Vijayaditya and Povey, Daniel and Seltzer, Michael L. and Khudanpur, Sanjeev},
}

@article{radford_robust_2022,
	title      = {Robust Speech Recognition via Large-Scale Weak Supervision},
	journal    = {arXiv preprint arXiv:2212.04356},
    year       = {2022},
	author     = {Radford, Alec and Kim, Jong Wook and Xu, Tao and Brockman, Greg and {McLeavey}, Christine and Sutskever, Ilya},
}

@inproceedings{roux_sdr_2019,
	title      = {{SDR} – Half-baked or Well Done?},
	year       = {2019},
	pages      = {626--630},
	booktitle  = {Proc. of the {IEEE} ICASSP},
	author     = {Roux, Jonathan Le and Wisdom, Scott and Erdogan, Hakan and Hershey, John R.},
}

@inproceedings{lagace_ego-noise_2023,
	title      = {Ego-Noise Reduction of a Mobile Robot Using Noise Spatial Covariance Matrix Learning and Minimum Variance Distortionless Response},
	year       = {2023},
	pages      = {3533--3538},
	booktitle  = {Proc. of the {IEEE}/{RSJ} IROS},
	author     = {Lagacé, Pierre-Olivier and Ferland, François and Grondin, François},
}

@article{ochiai_unified_2017,
	title      = {Unified Architecture for Multichannel End-to-End Speech Recognition With Neural Beamforming},
	volume     = {11},
    year       = {2017},
	pages      = {1274--1288},
	number     = {8},
	journal    = {{IEEE} JSTSP},
	author     = {Ochiai, Tsubasa and Watanabe, Shinji and Hori, Takaaki and Hershey, John R. and Xiao, Xiong},
}

@inproceedings{araki_comparison_2018,
  author    = {Shoko Araki and Nobutaka Ono and Keisuke Kinoshita and Marc Delcroix},
  title     = {Comparison of Reference Microphone Selection Algorithms for Distributed Microphone Array Based Speech Enhancement in Meeting Recognition Scenarios},
  booktitle = {Proc. of the IWAENC},
  pages     = {316--320},
  year      = {2018},
  doi       = {10.1109/IWAENC.2018.8521251}
}

@inproceedings{lawin-ore_reference_2012,
  author    = {Toby Christian Lawin-Ore and Simon Doclo},
  title     = {Reference Microphone Selection for {MWF}-Based Noise Reduction Using Distributed Microphone Arrays},
  booktitle = {Proc. 10th ITG Symp. Speech Communication},
  pages     = {1--4},
  year      = {2012}
}

@article{zhang_study_2021,
  author  = {Jie Zhang and Huawei Chen and Li-Rong Dai and Richard Christian Hendriks},
  title   = {A Study on Reference Microphone Selection for Multi-Microphone Speech Enhancement},
  journal = {IEEE/ACM Trans. Audio, Speech, Lang. Process.},
  volume  = {29},
  pages   = {671--683},
  year    = {2021},
  doi     = {10.1109/TASLP.2020.3039930}
}

@article{grimm_phase_2016,
  author  = {Simon Grimm and Toby Christian Lawin-Ore and Simon Doclo and J{\"u}rgen Freudenberger},
  title   = {Phase Reference for the Generalized Multichannel Wiener Filter},
  journal = {EURASIP J. Adv. Signal Process.},
  volume  = {2016},
  number  = {1},
  pages   = {78},
  year    = {2016},
  doi     = {10.1186/s13634-016-0375-6}
}

\end{document}